\documentclass{article}
\usepackage{spconf,amsmath,graphicx}
\usepackage{amsmath}
\usepackage{graphicx}
\usepackage{hyperref}
\usepackage{xspace}
\usepackage{arydshln}
\usepackage{mathtools}
\usepackage{threeparttable}
\usepackage{caption}
\usepackage{amsmath}
\usepackage{amssymb}
\usepackage{graphicx} 
\usepackage{booktabs}
\usepackage{multirow}
\usepackage{comment}
\usepackage{arydshln}
\usepackage{xcolor}
\usepackage{multicol}
\usepackage{tabularray}
\usepackage{tabularx}
\usepackage[most]{tcolorbox}
\usepackage{enumitem}
\usepackage{fancyhdr}
\usepackage{arydshln}

\usepackage{diagbox}

\copyrightnotice{\copyright IEEE 2026}
\toappear{
    \begin{tabular}[t]{@{}l@{}}
        \copyright This work has been submitted to the IEEE for possible publication.\\Copyright may be transferred without notice, after which this version may no longer be accessible.
   \end{tabular}
}

\definecolor{embed}{RGB}{153, 0, 255}
\definecolor{kernel}{RGB}{230, 230, 0}

\title{Seeing Roads Through Words: A Language-Guided Framework for RGB-T Driving Scene Segmentation}
 \name{Ruturaj Reddy$^{1,2{\star}}$, Hrishav Bakul Barua$^{1,3{\star}}$, Junn Yong Loo$^1$, Thanh Thi Nguyen$^{2,4}$, Ganesh Krishnasamy$^1$
 \thanks{$^{\star}$This research is supported by the Global Excellence and Mobility Scholarship (GEMS), Monash University (Australia \& Malaysia).}
 }
 \address{$^1$School of Information Technology, Monash University Malaysia, Malaysia\\
 $^2$Faculty of Information Technology, Monash University, Australia\\ 
 $^3$Robotics and Autonomous Systems Lab, TCS Research, Kolkata, India\\
 $^4$School of Science, Technology and Engineering,
University of the Sunshine Coast, Australia\\
\{ruturaj.reddy, hrishav.barua, loo.junnyong, ganesh.krishnasamy\}@monash.edu, tnguyen5@usc.edu.au}

\makeatletter
\DeclareRobustCommand\onedot{\futurelet\@let@token\@onedot}
\def\@onedot{\ifx\@let@token.\else.\null\fi\xspace}
\def\eg{\emph{e.g}\onedot} 
\def\ie{\emph{i.e}\onedot} 
 
\def\etc{\emph{etc}\onedot}

\makeatother
\begin{document}
%

\newcommand{\mymethod}{CLARITY}

\maketitle
\begin{abstract}
    Robust semantic segmentation of road scenes under adverse illumination, lighting, and shadow conditions remain a core challenge for autonomous driving applications. RGB-Thermal fusion is a standard approach, yet existing methods apply static fusion strategies uniformly across all conditions, allowing modality-specific noise to propagate throughout the network. Hence, we propose \mymethod~that dynamically adapts its fusion strategy to the detected scene condition. Guided by vision-language model (VLM) priors, the network learns to modulate each modality's contribution based on the illumination state while leveraging object embeddings for segmentation, rather than applying a fixed fusion policy. We further introduce two mechanisms - one which preserves valid dark-object semantics that prior noise-suppression methods incorrectly discard, and a hierarchical decoder that enforces structural consistency across scales to sharpen boundaries on thin objects. Experiments on the MFNet dataset demonstrate that \mymethod~establishes a new state-of-the-art (SOTA), achieving 62.3\% mIoU and 77.5\% mAcc.
\end{abstract}
\section{Introduction}
\label{sec:intro}

\begin{figure}[ht]
    \centering
    \includegraphics[width=1.0\linewidth]{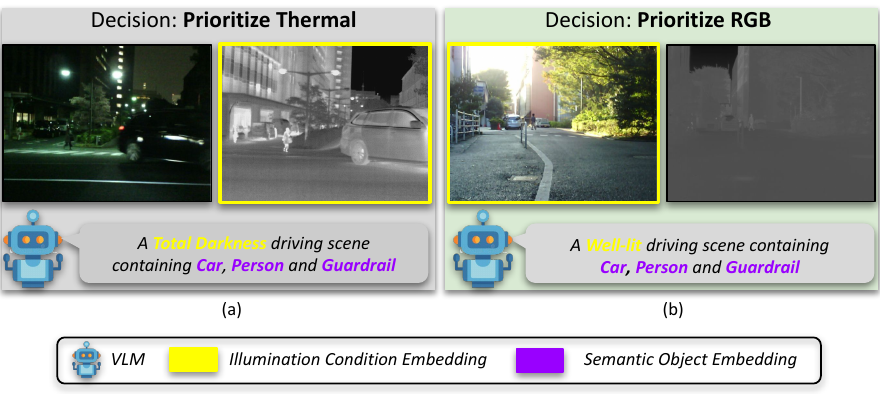}
    \caption{VLM generates a template text caption for each scene, which is encoded into a semantic object embedding (words in \textcolor{embed}{\textbf{violet}}) regardless of illumination conditions. (a) In poor illumination, the RGB image suffers from motion blur, obscuring the pedestrian. The thermal image remains clear. The VLM identifies this as a ``Total Darkness" condition (words in \textcolor{kernel}{\textbf{yellow}}), activating kernels (\textit{convolutional filters}) that prioritize thermal data to avoid integrating RGB noise. (b) Conversely, the RGB image provides clear details and edges for objects like the person and guardrail in ``Well-lit'' driving scene. The VLM gates the model to prioritize RGB over thermal input.}
    \label{fig:teaser}
\end{figure}

Semantic segmentation~\cite{7913730} is fundamental to robotic vision and autonomous driving safety~\cite{9525313}, enabling precise identification of navigation elements~\cite{Nesti_2022_WACV}. While RGB-Thermal (RGB-T) fusion~\cite{hazra2025cross,10234530} mitigates the illumination dependence of RGB-only methods, current frameworks face two limitations: \textbf{(a) Static fusion architectures:} Methods like CMX~\cite{10231003} and RTFNet~\cite{8666745} use fixed weights regardless of conditions, preventing aggressive suppression of noisy RGB branches (\eg, at night); and \textbf{(b) Feature divergence:} Hierarchical decoders often suffer from divergence where coarse and fine features fail to converge~\cite{11072373}, blurring thin objects.

To address this, we propose \mymethod, a Language-Guided expert framework. 
Inspired by Language-Driven Restoration (LDR)~\cite{10655208}, we utilize the Contrastive Language-Image Pre-training (CLIP)~\cite{radford2021learning} based vision-language model (VLM)~\cite{zhou2024vision} to generate illumination condition and class-specific semantic object embeddings. 
Unlike static networks, our fusion module uses the illumination condition embedding to dynamically activate only the most relevant ``Expert'' kernel (\eg, a ``Night Expert'' prioritizing thermal features), while semantic object embedding serves as semantic priors for ambiguous targets (see Fig.~\ref{fig:teaser}). We also transfer the Self-Calibrated Illumination (SCI) framework~\cite{11072373} into the decoder to enforce consistency. 

Early RGB-T approaches such as MFNet~\cite{ha2017mfnet} and recent Transformer-based methods such as MMSMCNet~\cite{10123009} remain ``condition-agnostic,'' applying uniform fusion logic regardless of severity. We advance this by introducing condition-specific dynamic gating inspired by the Mixture-of-Experts (MoE) approach. Recent works, such as CAFuser~\cite{10858375}, use text tokens to guide fusion. We extend this paradigm to RGB-T segmentation, using embeddings to dynamically gate the fusion and teaching the network to prioritize thermal features when prompts indicate conditions such as ``Total Darkness.'' Unlike static layers, this semantic gating recalibrates feature importance layer-by-layer, preventing noise contamination during critical safety scenarios. Novel priors from low-light enhancement remain underutilized. We address this by adapting SCI~\cite{11072373} to eliminate structural artifacts in objects. We use ``unbalanced point transformer prior'' from UPT-Flow~\cite{XU2025111076} to flag distorted RGB pixels, ensuring robustness against illumination variations. Our \textbf{key-contributions} are as follows:

\begin{itemize}
    \item \textbf{Language-Guided Dynamic Fusion}: We propose an RGB-T segmentation network that utilizes CLIP-derived semantic prompts to gate a Sparse MoE~\cite{10655208} based on illumination conditions, while simultaneously integrating object embeddings to guide segmentation.
    \item \textbf{Soft-Gated Unbalanced Attention}: We introduce a ``Soft'' unbalanced point mechanism~\cite{XU2025111076} that identifies distorted RGB pixels via variance mapping. Unlike hard-thresholding methods that blind the model to dark objects, our soft-gating preserves critical color cues (\eg, Traffic Cones) while suppressing noise. 
    \item \textbf{Self-Calibrated Decoding}: We design the self-calibration module based on illumination learning~\cite{11072373} to bridge the decoder stages, significantly reducing boundary artifacts and achieving SOTA on overall metrics.
\end{itemize}

\section{Method}
\subsection{Architectural Overview of \mymethod}
\begin{figure*}[ht]
    \centering
    \includegraphics[width=.78\linewidth]{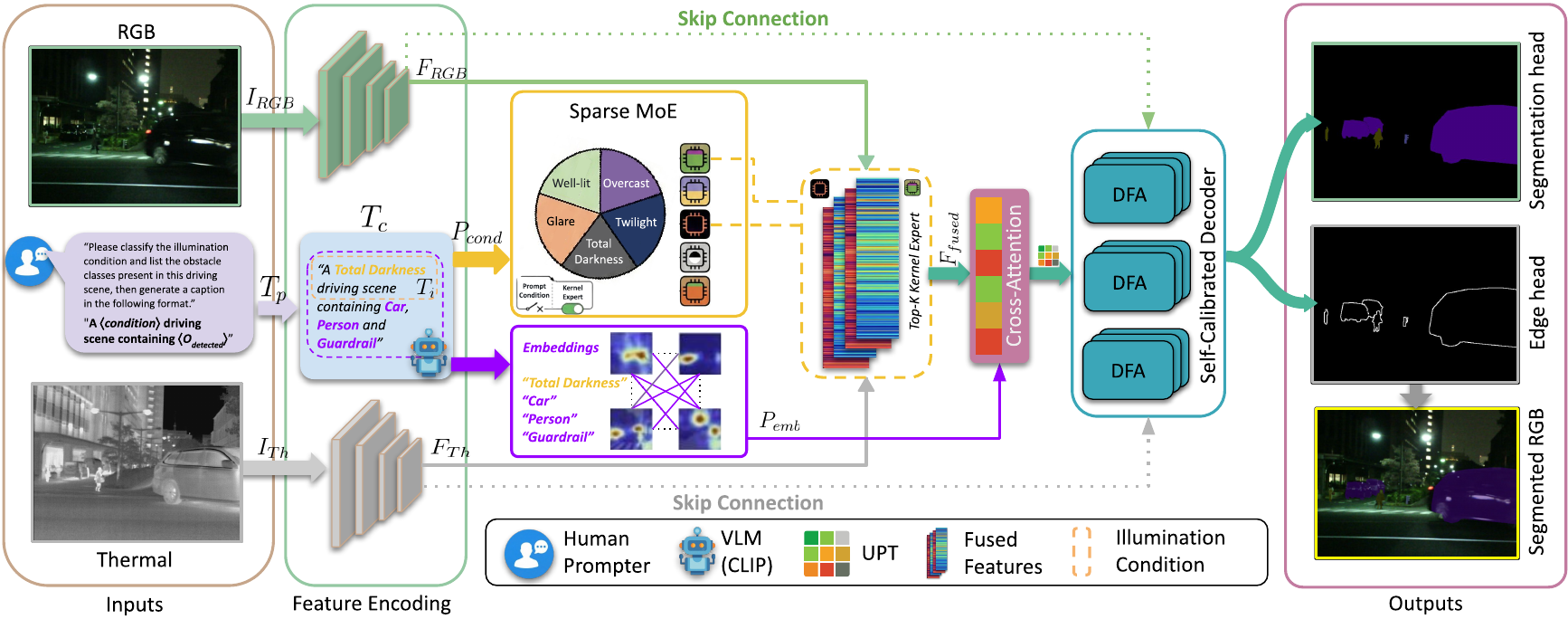}
    \caption{Architecture of the proposed \mymethod~method. The framework begins with a Semantic Condition Generator, where a VLM (CLIP) analyzes the scene to generate caption $T_c$. This process yields two key embeddings, \ie, an illumination condition embedding ($P_{cond}$) from $T_i$ derived for illumination conditions ($\textcolor{kernel}{\langle condition\rangle}$) and a holistic scene embedding ($P_{emb}$) derived from $T_c$ for detected objects ($\textcolor{embed}{\langle O_{detected}\rangle}$) and illumination condition. $P_{cond}$ serves as a gating signal for the Sparse MoE, dynamically routing inputs to degradation-specific \textit{Top-K} Kernel Experts for adaptive fusion. The fused features are then enhanced by a Soft-Gated UPT to recover faint thermal details. Finally, a Self-Calibrated Decoder with Dilated Feature Aggregation (DFA) blocks captures multi-scale context before feeding the Segmentation and Edge Heads.}
    \label{fig:architecture}
\end{figure*}

The proposed framework for training (Fig.~\ref{fig:architecture}) adopts a dual-stream structure via three restoration-based mechanisms for fusion. Specifically, the encoder integrates a Language-Guided MoE that utilizes semantic prompts to gate glare-affected pixels to thermal experts and inject object embeddings. This is complemented by a Soft-Gated Unbalanced Point Transformer (SG-UPT) to preserve faint details in low-confidence regions, while a Self-Calibrated Decoder enforces inter-stage feature consistency to prevent modality conflict.

\subsection{Feature Extraction and Semantic Conditioning}
 VLMs are becoming increasingly popular in guiding low-level vision tasks such as Language-Driven Restoration (LDR)~\cite{10655208}. In their work, CLIP detects weather degradations to activate specific noise-removal kernels. We use CLIP-based VLM to extract a \textit{Illumination Condition Embedding} $P_{cond}$ for gating the fusion process and a \textit{Holistic Scene Embedding} (consisting of both illumination condition and semantic object embeddings) $P_{emb}$ for semantic scene understanding. Given an RGB image $I_{RGB} \in \mathbb{R}^{H \times W \times 3}$ and a thermal image $I_{Th} \in \mathbb{R}^{H \times W \times 1}$, we finetune two pretrained ConvNeXt~\cite{Liu_2022_CVPR} networks to extract hierarchical feature maps $\{F^i_{RGB}, F^i_{Th}\}_{i=1}^4$. Simultaneously, to capture the nuance of the driving environment, we generate a comprehensive prompt $T_p$ given an input $I_{RGB}$ to interrogate the scene context. Crucially, to ensure strict semantic alignment with the downstream segmentation task, we define fixed constraints for both illumination conditions and objects detected within the prompt. First, we establish a set of distinct illumination conditions $C_{cond}$ consisting of: $\langle$\textit{Glare, Well-lit, Overcast, Twilight,} and \textit{Total Darkness}$\rangle$. Second, we constrain object descriptors to the pre-defined MFNet~\cite{8206396} class set $C_{target}$, including: $\langle$\textit{Car, Person, Bike, Curve, Car Stop, Guardrail, Color Cone,} and \textit{Bump}$\rangle$. The prompt $T_p$ is therefore constructed textually as: 
\begin{equation}
\scriptsize
\text{``A }
\underbrace{\langle \text{{\textit{condition}}} \rangle}_{\mathclap{\{\text{\textbf{\textcolor{kernel}{Glare}, \textcolor{kernel}{Well-lit}, \textcolor{kernel}{Overcast}, \etc}}\}}}
\text{ driving scene containing }
\underbrace{\langle O_{{\textit{detected}}} \rangle}_{\mathclap{\{\text{\textbf{\textcolor{embed}{Car}, \textcolor{embed}{Person}, \textcolor{embed}{Bike}, \etc}}\}}}
\text{''},
\end{equation}
where, the chosen $\langle \text{\textit{condition}} \rangle \in C_{cond}$, and the detected objects $\langle O_{detected} \rangle \subseteq C_{target}$ (an example caption $T_c$ generated - ``\textit{A \textbf{\textcolor{kernel}{Total Darkness}} driving scene containing \textbf{\textcolor{embed}{Car}}, \textbf{\textcolor{embed}{Person}} and \textbf{\textcolor{embed}{Guardrail}}}'', see Figs.~\ref{fig:teaser} and ~\ref{fig:architecture}). By filtering the VLM output to strictly match these targeted sets, we not only prevent open-vocabulary noise from corrupting the condition vector but also gain a significant advantage in detecting challenging objects. Classes such as `Guardrail' are often visually ambiguous or obscure in MFNet images due to poor illumination. By explicitly querying CLIP based on its vast pre-trained semantic knowledge, we inject a strong, task-specific prior into the system. This ensures the fusion process is heavily biased towards features relevant to these hard-to-find classes, rather than just dominant background elements.

The caption $T_c$ begins with the illumination condition phrase $T_i$ (\eg, ``\textit{A \textbf{\textcolor{kernel}{Total Darkness}} driving scene}''). Acting as a Task-Aligned Global Controller, $T_i$ directs the MoE fusion to activate \textit{Top-K} experts (\textit{K=2}). We encode $T_i$ ($P_{cond}$) using a frozen CLIP encoder followed by a Multilayer Perceptron (MLP) for alignment. Similarly, the complete caption $T_c$ is processed through the same pipeline to yield $P_{emb}$, an embedding that encodes both the illumination condition and specific semantic classes (\eg, Guardrail):
    \begin{equation}
    \scriptsize
    P_{cond} = \text{MLP}(\text{CLIP}_{enc}(T_i)), \quad P_{cond} \in \mathbb{R}^{C},
    \end{equation}
    \begin{equation}
    \scriptsize
    P_{emb} = \text{MLP}(\text{CLIP}_{enc}(T_c)), \quad P_{emb} \in \mathbb{R}^{C},
    \end{equation}
where $C$ denotes the channel dimensionality of the encoder feature maps. 

\subsection{Soft-Gated Unbalanced Point Attention (SG-UPT)}
The Unbalanced Point Prior (UPT)~\cite{XU2025111076} is originally introduced for low-light image enhancement, employing a hard threshold ($T=0.1$) on RGB ratios to binary mask out pixels with severe color distortion. We identify that such hard-thresholding is detrimental to segmentation as it creates binary gradient blockers that sever the semantic signal of valid dark objects (\eg, traffic cones). Therefore, we propose a \textit{Soft-Gated Tanh Bias}. First, we design the Unbalanced Map $M_{un} \in \mathbb{R}^{H \times W}$ to capture the specific artifacts of driving scenes (e.g., noise and underexposure) rather than the color ratio. We formulate $M_{un}$ based on local variance and intensity:
\begin{equation}
\scriptsize
 M_{un} = \sigma_{\text{local}}(I_{RGB}) + (1 - \mu_{\text{local}}(I_{RGB})),
\end{equation}
where $\sigma_{\text{local}}$ and $\mu_{\text{local}}$ denote local variance and mean pooling operations, respectively. This map highlights regions with high noise (high variance) or extreme darkness (low mean).

Second, we inject this map into the attention mechanism as a soft bias rather than a hard mask. Let $Q, K, V \in \mathbb{R}^{N \times d}$ be the projections of the flattened feature map $X \in \mathbb{R}^{N \times C}$ (where $N=H \times W$). The refined feature $\hat{X}$ is computed as:
\begin{equation}
\scriptsize
\hat{X} = \text{Softmax}\left(\frac{QK^T}{\sqrt{d}} - \lambda \cdot \tanh(M_{un}^{\text{flat}})\right) V,
\end{equation}
where $M_{un}^{\text{flat}} \in \mathbb{R}^{1 \times N}$ is the flattened and broadcasted map, $\lambda$ is a learnable scalar, and $d$ is the channel dimension of the query/key projections, serving as a scaling factor to prevent gradient saturation in the Softmax operation. The $\tanh$ term softly down-weights the attention scores for noisy or glare-affected pixels (where $M_{un}$ is high), effectively suppressing their contribution to the global context without completely discarding their features.

\begin{table*}[ht]
\small
\centering

\caption{Comparison of semantic segmentation results on the MFNet dataset~\cite{10234530}. Best results are in \textbf{bold}.}
\label{tab:results}
\resizebox{.8\textwidth}{!}{
\begin{tabular}{l|cc|cc|cc|cc|cc|cc|cc|cc|cc}
\toprule
\multirow{2}{*}{\textbf{Methods}} & \multicolumn{2}{c|}{\textbf{Car}} & \multicolumn{2}{c|}{\textbf{Person}} & \multicolumn{2}{c|}{\textbf{Bike}} & \multicolumn{2}{c|}{\textbf{Curve}} & \multicolumn{2}{c|}{\textbf{Carstop}} & \multicolumn{2}{c|}{\textbf{Guardrail}} & \multicolumn{2}{c|}{\textbf{Cone}} & \multicolumn{2}{c|}{\textbf{Bump}} & \multirow{2}{*}{\textbf{mAcc}(\%)} & \multirow{2}{*}{\textbf{mIoU}(\%)} \\
& Acc & IoU & Acc & IoU & Acc & IoU & Acc & IoU & Acc & IoU & Acc & IoU & Acc & IoU & Acc & IoU & & \\
\midrule
FuseSeg (\textit{TASE'21})~\cite{9108585}
 & 93.1 & 87.9 & 81.4 & 71.7 & 78.5 & 64.6 & 68.4 & 44.8 & 29.1 & 22.7 & 63.7 & 6.4 & 55.8 & 46.9 & 66.4 & 47.9 & 70.6 & 54.5 \\
GMNet (\textit{TIP'21})~\cite{9531449} & 94.1 & 86.5 & 83 & 73.1 & 76.9 & 61.7 & 59.7 & 44 & 55 & 42.3 & \textbf{71.2} & 14.5 & 54.7 & 48.7 & 73.1 & 47.4 & 74.1 & 57.3 \\
MFNet (\textit{IROS'17})~\cite{ha2017mfnet} & 95.1 & 87.9 & 85.2 & 66.8 & 83.9 & 64.4 & 64.3 & 47.1 & 50.8 & 36.1 & 45.9 & 8.4 & 62.8 & 55.5 & 73.8 & 62.2 & 74.7 & 57.3 \\
RTFNet (\textit{RAL'19})~\cite{8666745} & 93.0 & 87.4 & 79.3 & 70.3 & 76.8 & 62.7 & 60.7 & 45.3 & 38.5 & 29.8 & 0.0 & 0.0 & 45.5 & 29.1 & 74.7 & 55.7 & 63.1 & 53.2 \\
DooDLeNet (\textit{CVPRW'22})~\cite{Frigo_2022_CVPR}& 91.7 & 86.7 & 81.3 & 72.2 & 76.6 & 62.5 & 58.9 & 46.7 & 36.2 & 28 & 35.2 & 5.1 & 56.9 & 50.7 & 74.8 & \textbf{65.8} & 67.9 & 57.3 \\
Parameter Sharing (\textit{TITS'24})~\cite{10337777} & 94.3 & 88.3 & 82.8 & 72.5 & 82.9 & 65.2 & 74.9 & 46.6 & 62.7 & \textbf{47} & 42.7 & 11.6 & 55 & 46 & 76 & 57.1 & 74.4 & 59.1 \\
EGFNet \textit{(TITS'24)}~\cite{10234530} & 95.8 & 87.6 & 89.0 & 69.8 & 80.6 & 58.8 & 71.5 & 42.8 & 48.7 & 33.8 & 33.6 & 7.0 & 65.3 & 48.3 & 71.1 & 47.1 & 72.7 & 54.8 \\
IGFNet (\textit{ROBIO'23})~\cite{10354613} & 93.2 & 88.0 & 83.4 & 74.0 & 71.8 & 62.7 & 67.6 & 48.2 & 45.4 & 36.0 & 68.5 & 14.2 & 58.8 & 52.4 & 68.3 & 57.5 & 72.9 & 59.0 \\
MMSMCNet (\textit{TCSVT'23})~\cite{10123009} & 96.2 & 89.2 & \textbf{93.2} & 69.1 & 83.4 & 63.5 & 74.4 & 46.4 & 56.6 & 41.9 & 26.9 & 8.8 & \textbf{70.2} & 48.8 & 77.5 & 57.6 & 75.2 & 58.1 \\
CMX(MiT-B2)$^*$ (\textit{TITS'23})~\cite{10231003} & 93.8 & 89.4 & 84.6 & 74.8 & 75.6 & 64.7 & 64.4 & 47.3 & 42.5 & 30.1 & 54.8 & 8.1 & 54.1 & 52.4 & 72.2 & 59.4 & 71.29 & 58.2 \\
CMX(MiT-B4) (\textit{TITS'23})~\cite{10231003} & - & 90.1 & - & \textbf{75.2} & - & 64.5 & - & \textbf{50.2} & - & 35.3 & - & 8.5 & - & 54.2 & - & 60.6 & - & 59.7 \\
CMAAHF (\textit{ICIP'25})~\cite{hazra2025cross} & 96.3 & 91.2 & 83.7 & 74.1 & 82.6 & 66.6 & 73.8 & 46.1 & 64.1 & 43.8 & 47.1 & 18.2 & 67.1 & \textbf{56.1} & \textbf{78.9} & 60.7 & 76.3 & 61.9 \\
\midrule
\textbf{\mymethod~(Ours)} & \textbf{97.4} & \textbf{91.5} & 86.0 & 74.8 & \textbf{84.8} & \textbf{67.2} & \textbf{76.2} & 46.6 & \textbf{66.3} & 44.3 & 49.3 & \textbf{18.7} & 66.2 & 55.7 & 78.2 & 60.1 & \textbf{77.5} & \textbf{62.3} \\
\bottomrule
\end{tabular}
}
\caption*{\footnotesize $^*$Reproduced using model checkpoint.}
\end{table*}

\subsection{Language-Guided Sparse MoE Fusion}
To achieve illumination condition-specific modality integration, we propose a Language-Guided Sparse (MoE) for dynamic feature fusion. Drawing inspiration from the degradation-adaptive routing in LDR~\cite{10655208}, we replace standard static fusion layers with a routable expert bank that reconfigures based on semantic context. We maintain $N$ convolution filters $E = \{E_1, ..., E_N\}$ and denote them as expert kernels. A gating network $W_g$ projects the condition vector embedding $P_{cond}$ and input features into a selection score $S$:
\begin{equation} 
    \scriptsize
    S = \text{Softmax}\left( W_g \left( [F_{RGB}, F_{Th}, \text{Expand}(P_{cond})] \right) \right), 
\end{equation}
where Expand($\cdot$) broadcasts the global condition vector $P_{cond}$ to the spatial dimensions H×W and $[\cdot]$ denotes concatenation and $S \in \mathbb{R}^{H \times W \times N}$ represents the gating weights. The fused feature $F_{fused}$ is the weighted summation of the Top-$K$ experts applied to the concatenated input features. Let $\rho(k)$ denote the index of the $k^{th}$ highest scoring expert at a given pixel. The sparse aggregation is formulated as:
\begin{equation}
\scriptsize
F_{fused} = \sum_{k=1}^{K} S_{\rho(k)} \odot E_{\rho(k)}([F_{RGB}, F_{Th}]),
\end{equation}
where $S_{\rho(k)} \in \mathbb{R}^{H \times W}$ is the spatial confidence map for the chosen expert $E_{\rho(k)}$. This mechanism allows the network to physically reconfigure per-pixel, activating the ``Thermal Expert" only at specific coordinates where glare or darkness is detected. Finally, we employ a standard cross-attention between $F_{fused}$ and $P_{emb}$ to inject holistic semantic scene context into the spatial features.

\subsection{Self-Calibrated Hierarchical Decoder}
Inspired by the Self-Calibrated Illumination (SCI) framework~\cite{11072373}, which enforces exposure stability via weight sharing, we design a task-specific variant to alleviate feature divergence in hierarchical decoding. In standard decoders, error accumulation across upsampling stages often blurs the boundaries of small objects. To prevent this, we implant a self-calibrator to anchor the decoding trajectory.

Let $D_t$ be the decoder feature at stage $t$. We utilize a parameter-sharing decoding block $\mathcal{H}_\theta$ instantiated as a Dilated Feature Aggregation (DFA) module. The DFA block~\cite{hazra2025cross} applies parallel convolutions with varying dilation rates to capture multi-scale context without resolution loss. Drawing on the convergence property of SCI, we force the output of the subsequent stage $D_{t+1}$ to be calibrated against the initial structural representation $D_1$:
\begin{equation}
\scriptsize
 D_{t+1} = \mathcal{H}_{\theta}\left( \text{Up}(D_t) + \mathcal{S}_{\phi}(D_1) \right),
\end{equation}
\noindent where $\text{Up}(\cdot)$ denotes bilinear upsampling and $\mathcal{S}_{\phi}$ acts as a learnable calibrator. This formulation ensures that the DFA blocks remain centered around the fixed initial estimate $D_1$ rather than drifting due to cascading dependencies. Here, $\mathcal{S}_{\phi}$ acts as a learnable calibrator that projects the coarse semantic anchor $D_1$ to match the resolution of stage $t$. This explicitly suppresses the ``error clusters'' around high-frequency edges, allowing multi-scale DFA features to sharpen the boundaries of thin structures such as guardrails.

\subsection{Edge-Aware Loss Function}
To resolve the severe class imbalance between edge and non-edge pixels, we employ a joint loss combining Weighted Cross-Entropy ($\mathcal{L}_{seg}$) for semantic segmentation and Focal Loss ($\mathcal{L}_{edge}$) for the auxiliary edge head. The total objective is formulated as:
\begin{equation}
\scriptsize
\mathcal{L}_{total} = \mathcal{L}_{seg}(P, G) + \beta \mathcal{L}_{edge}(P_e, G_e),
\end{equation}
where $P, G$ and $P_e, G_e$ denote the semantic and edge predictions and ground truths, respectively. We empirically set $\beta=0.6$ to balance the auxiliary task gradient.

\section{Experiments and Results}
\textbf{Implementation:} We implement our method using PyTorch, trained on an Nvidia A40 GPU (48GB VRAM), utilizing the AdamW optimizer and a poly learning rate schedule.

\noindent \textbf{Datasets:} We evaluate on the popular MFNet dataset~\cite{10234530}, comprising 1,569 aligned RGB-T pairs ($480 \times 640$) captured in urban scenes via an \textit{InfReC} \textit{R500} camera. Following standard protocols~\cite{8666745}, the data is split into 784 training, 392 validation, and 393 testing images. The dataset annotates nine classes: one background and eight obstacles (\eg, \textit{Car, Guardrail}). A critical challenge is the severe ``long-tail'' imbalance where objects like \textit{Color Cones} are rare, necessitating our weighted loss formulation. Furthermore, while the metadata distinguishes 820 daytime and 749 nighttime images, we find these binary labels insufficient to capture sensor degradations, motivating the detailed categorization in methodology.

\noindent \textbf{Methods:} To validate the superiority of our method, we benchmark against representative state-of-the-art methods. We compare with CNN-based fusion networks including MFNet~\cite{ha2017mfnet}, RTFNet~\cite{8666745}, and EGFNet~\cite{10234530}, which utilize dual-stream encoders. Additionally, we compare against recent Transformer-based approaches such as CMX~\cite{10231003}, which employs cross-modal fusion, and CMAAHF~\cite{hazra2025cross}, which utilizes hierarchical attention.

\begin{figure}
    \centering
    \includegraphics[width=1.0\linewidth]{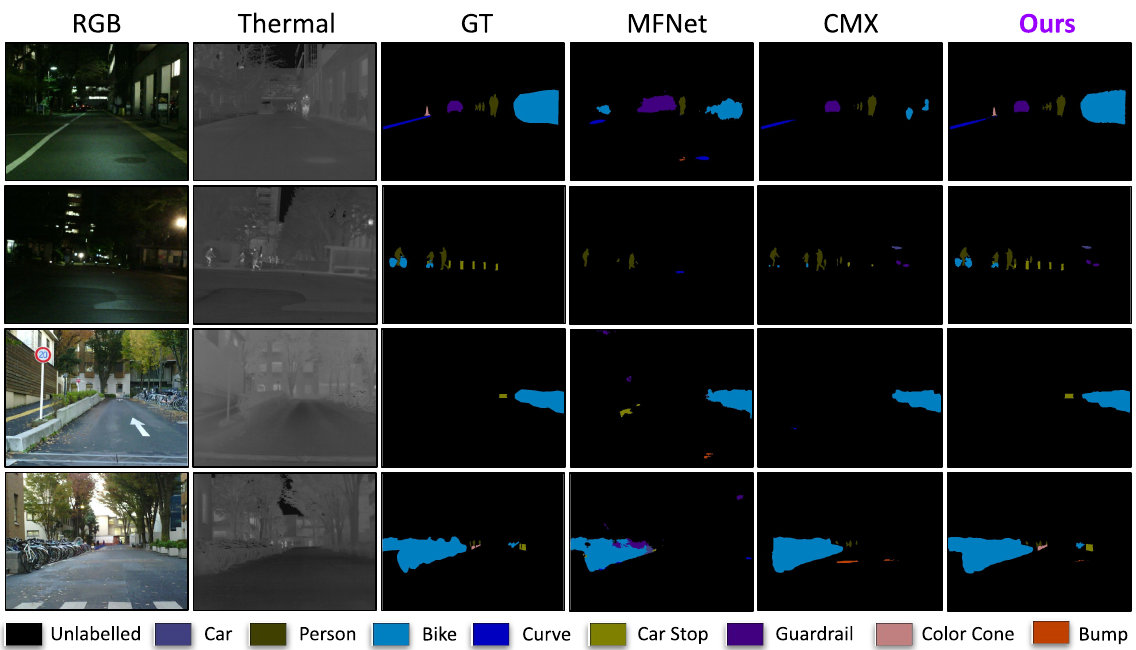}
    \caption{Segmented results generated by the proposed \mymethod~and state-of-the-art methods on MFNet dataset~\cite{10234530}.}
    \label{fig:qualitative_results}
\end{figure}

\noindent\textbf{Quantitative Results.} Table~\ref{tab:results} compares performance on the MFNet test set~\cite{10234530}. While the CMX baseline fragments thin structures like \textit{Guardrails} (8.1\% IoU) and misses \textit{Color Cones} due to weak thermal signatures, \mymethod~utilizes Soft-Gated UPT to explicitly recover these details, boosting \textit{Guardrail} IoU to 18.7\%. Our CLIP-generated prompts provide semantic priors that enable the detection of objects barely visible in both modalities, yielding continuous masks and raising \textit{Car} IoU to 91.5\% compared to 89.4\% for CMX. \mymethod~sets a new state-of-the-art with 62.3\% mIoU and 77.5\% mAcc, surpassing CMAAHF (61.9\% mIoU). This performance advantage extends to dynamic classes such as \textit{Bike} (67.2\% IoU), confirming that our semantic gating effectively handles both static structural elements and moving obstacles better than standard fusion baselines.

\begin{table}[h]
\centering
\caption{Component Contribution Analysis on the MFNet dataset. We evaluate the incremental impact of the Sparse Mixture-of-Experts (MoE), the Language-Driven Gating and scene embedding via CLIP, and the Soft-Gated UPT (SG-UPT) on the baseline Dual-Stream architecture.}
\label{tab:ablation_modules}
\resizebox{0.5\columnwidth}{!}{ 
\begin{tabular}{l|l}
\toprule
 \textbf{Model Configuration} & \textbf{mIoU (\%)}\\
\midrule
 Baseline (Dual ConvNeXt~\cite{Liu_2022_CVPR}) & 56.5 \\
 + Sparse MoE (Random Router) & 57.7~\textit{(+1.2)} \\
 + $CLIP_{P_{cond}}$ & 60.1~\textit{(+2.4)} \\
 + $CLIP_{P_{emb}}$ & 61.0~\textit{(+0.9)} \\
 + Soft-Gated UPT & \textbf{62.3}~\textit{(+1.3)} \\
\bottomrule
\end{tabular}
}
\end{table}

\noindent\textbf{Qualitative Results:}
Fig.~\ref{fig:qualitative_results} presents a qualitative comparison on the MFNet test set~\cite{10234530}. While CMX provides a strong baseline, it exhibits structural fragmentation on thin objects (\eg, guardrails) and frequently fails to detect Color Cones, often misclassifying them as background due to their limited thermal footprint. In contrast, \mymethod~utilizes the Soft-Gated UPT via VLM gating to explicitly recover these faint details. Notably, the integration of CLIP-generated prompts enables our model to detect objects barely visible in both RGB and thermal modalities by leveraging semantic priors (\textit{embeddings}) where visual cues are insufficient. This yields significantly sharper boundaries and complete masks compared to the coarser, fragmented predictions of MFNet.

\begin{table}[b]
    \centering
    \caption{Comparison of Attention Gating Mechanisms. We compare the original UPT hard thresholding ($T=0.1$) against our proposed Soft Tanh modulation.}
    \label{tab:ablation_upt}
    \resizebox{0.6\columnwidth}{!}{
    \begin{tabular}{l|c|l}
        \toprule
        \textbf{Mechanism} & \textbf{Gating Function} &  \textbf{mIoU(\%)} \\
        \midrule
        No Prior & - & 60.1 \\
        Original UPT~\cite{XU2025111076} & Hard Cut ($P < 0.1$) & 60.5~\textit{(+0.4)} \\
        SG-UPT (Ours) & Soft Tanh ($\tanh(\alpha P)$) & \textbf{62.3}~\textit{(+1.8)} \\
        \bottomrule
    \end{tabular}
    }
\end{table}

\noindent\textbf{Ablation Study:}
We investigate the impact of the MoE architecture and the Language-Driven Gating and embedding. As shown in Table~\ref{tab:ablation_modules}, replacing the standard decoder with a ``Random Router" MoE provides only marginal gains (+1.2\%). However, enabling CLIP Gating via $P_{cond}$, which allows the model to select experts based on the prompt, yields a significant boost to 60.1\% mIoU. Then the introduction of CLIP-generated scene embedding $P_{emb}$ increases the performance further to 61.0\%. The final addition of the Soft-Gated UPT pushes performance to 62.3\%, confirming that semantic routing and detailed feature preservation are complementary.

We validate our modification to the Unbalanced Point Transformer (UPT). The original UPT uses a hard threshold (T=0.1) to filter distorted pixels. As shown in Table~\ref{tab:ablation_upt}, this hard cut suppresses faint objects like Guardrails (13.1\% IoU). Our proposed Soft Tanh gating preserves gradients for these low-light pixels, improving Guardrail recovery by over 5\%.

\begin{table}[t]
\centering
\caption{Impact of Semantic Prompt Granularity. Fine-grained illumination conditions allow experts to specialize in specific degradations as compared to binary labels only.}
    \label{tab:ablation_prompts}
    \resizebox{.9\columnwidth}{!}{
    \begin{tabular}{l|l|l}
        \toprule
        \textbf{Prompt Strategy} & \textbf{Condition Set ($C_{cond}$)} & \textbf{mIoU (\%)} \\
        \midrule
        No Prompt (w/o VLM) & - & 57.7 \\
        Binary Labels & \{Day, Night\} & 59.2~\textit{(+1.5)} \\
        Ternary Labels & \{Day, Overcast, Night\} & 60.5~\textit{(+1.3)} \\
        Fine-Grained (Ours) & \{Glare, Well-lit, Overcast, Twilight, Total Darkness\} & \textbf{62.3}~\textit{(+1.8)} \\
        \bottomrule
    \end{tabular}
    }
\end{table}

Finally, we analyze prompt engineering for the gating mechanism. Table~\ref{tab:ablation_prompts} demonstrates that using the dataset's native binary labels (``Day", ``Night") is insufficient (59.2\% mIoU) because ``Night" conflates distinct degradation modes like pitch darkness and headlight glare. Similarly, we observe only a minute improvement with ternary labels (``Day", ``Overcast", ``Night"). Finally, our fine-grained 5-condition stratum allows the MoE to specialize kernels for these specific noise distributions, achieving optimal results.

\section{Conclusions}
In this work, we introduce a dynamic fusion framework designed to address the vulnerability of static RGB-T networks under varying illumination conditions. By integrating a semantic gating mechanism driven by fine-grained prompts, our model dynamically routes pixels to degradation-specific experts, significantly outperforming coarser binary or ternary conditioning strategies. Furthermore, the proposed SG-UPT resolves the feature-suppression issue inherent to hard-thresholding priors, enabling recovery of faint, safety-critical objects such as guardrails and color cones in low-light scenarios. Extensive evaluations on the MFNet benchmark demonstrate that our approach establishes a new state-of-the-art with 62.3\% mIoU and 77.5\% mAcc, outperforming recent Transformer-based methods such as CMX and CMAAHF. These results confirm that injecting semantic condition awareness into the fusion process is a key factor for robust autonomous perception. Future work will focus on extending this approach to adverse weather scenarios (\ie, foggy, rainy, snowy, \etc) to enhance the diversity of kernel experts via VLM gating.

\newpage
\bibliographystyle{IEEEbib}
{\small 
 \bibliography{strings,refs}
}

\end{document}